# Joint Object-Material Category Segmentation from Audio-Visual Cues

Anurag Arnab[1]
aarnab@robots.ox.ac.uk

Michael Sapienza[1]
michael.sapienza@eng.ox.ac.uk

Stuart Golodetz[1]
stuart.golodetz@ndcn.ox.ac.uk

Julien Valentin[1]
julien.valentin@eng.ox.ac.uk

Ondrej Miksik[1]
ondrej.miksik@eng.ox.ac.uk

Shahram Izadi[2]
shahrami@microsoft.com

Philip H. S. Torr[1]
philip.torr@eng.ox.ac.uk

[1] Department of Engineering Science
University of Oxford
Oxford, UK

[2] Microsoft Research
Redmond, US

This research was supported by Technicolor, the EPSRC, the Leverhulme Trust and the ERC grant ERC-2012-AdG 21162-HELIOS.

## Abstract

It is not always possible to recognise objects and infer material properties for a scene from visual cues alone, since objects can look visually similar whilst being made of very different materials. In this paper, we therefore present an approach that augments the available dense visual cues with sparse auditory cues in order to estimate dense object and material labels. Since estimates of object class and material properties are mutually-informative, we optimise our multi-output labelling jointly using a random-field framework. We evaluate our system on a new dataset with paired visual and auditory data that we make publicly available. We demonstrate that this joint estimation of object and material labels significantly outperforms the estimation of either category in isolation.

## 1 Introduction

The ability to segment visual data into semantically-meaningful regions is vitally important in many computer vision scenarios, including automatic scene description [37], autonomous robot navigation [35], grasping [4] and assisted navigation for the visually-impaired [25]. Although existing semantic segmentation methods [5, 13, 17, 22, 34, 38] have achieved impressive results, their reliance on visual features makes it hard for them to distinguish between distinct object classes that are visually similar, even when the classes have different material properties.

An example of this can be seen in Fig. 1(a), in which the table, wall and mug all look very similar in terms of local colour and texture, even though the table is made from wood, the wall from gypsum and the mug from ceramic. In this case, knowing the material properties of the different entities would help in disambiguating between them, but such knowledge can be difficult or impossible to acquire from visual data alone. One way of resolving this problem is to make use of additional sensory modalities to complement the available visual information. For example, Miksik *et al*. [24, 25] used an active laser to help a passive camera resolve ambiguities in dense matching and depth estimation.

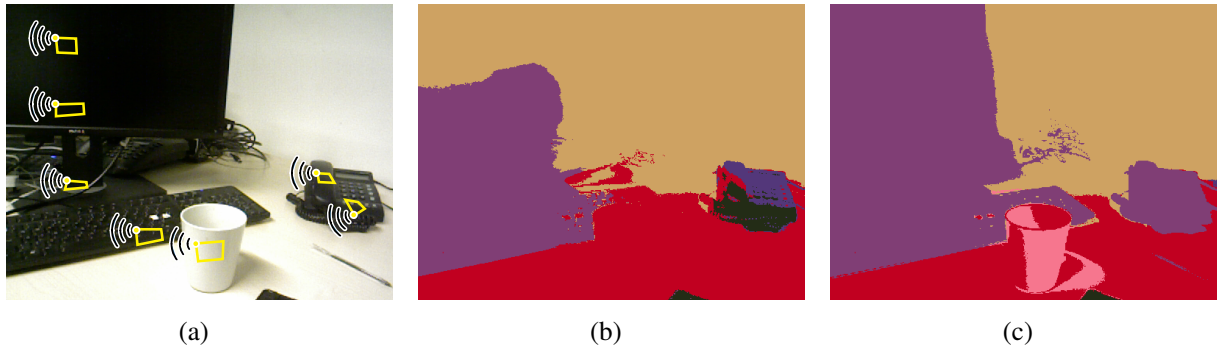

Figure 1: *(a) An image of a typical office desk. Notice the minimal amount of texture or colour present in the image. The yellow boxes indicate places in which we tap objects to produce audio features. (b) The results of predicting materials for the scene using only visual features (red = wood, purple = plastic, pink = ceramic, gold = gypsum). (c) The improved predictions obtained when we additionally incorporate audio features. The improvements for the monitor, keyboard and phone are especially noticeable.*

The problem of determining the material properties of objects based only on their visual appearance is difficult not just for computers, but also for human beings. One way in which we tend to resolve this problem in real life is to make use of the sounds that an object makes when we tap it to infer its properties. In this paper, we therefore present an approach that uses sound to augment the available dense visual cues with sparse auditory cues in order to estimate material labels, and then uses this information to improve object labelling as well. This approach is inspired by the McGurk effect – a perceptual phenomenon which shows that humans process auditory and visual information jointly when interpreting speech [18].

The difference that this makes to material labelling can be seen in Fig. 1. In (b), we see the results of performing a visual-only segmentation to produce material labels for the scene in (a): observe how it has failed to predict correct material labels for the mug and telephone, while also making errors on the monitor and keyboard. By contrast, when we tap various objects in the scene (a) to produce audio features that we can incorporate into our segmentation process, the predicted material labels significantly improve (c). These better material predictions are not only intrinsically useful, but also help us achieve better object-class predictions (Section 5.1).

Existing object segmentation datasets [6, 7, 13, 30, 31, 36] do not provide audio-visual annotations as ground truth. Furthermore, it is not possible to simply augment them with audio data, since we would need the original objects in the dataset in order to extract sound. For these reasons, we designed our own indoor dataset (Section 3) with dense per-pixel object category and material labels, which we make publicly available. We demonstrate the effectiveness of our approach on this new dataset by showing that it produces semantic segmentation results that are significantly better than those obtained when using visual features alone.

## 2 Background

Sinapov *et al*. [32] investigated using sound emitted from an object to predict its object category, classifying small objects based on sounds generated when interacting with a robotic arm (including the sound of the robot's motors). Harrison [15] observed that the sound emitted from an object may be used to predict its material category when exploring three objects (hand, paper, LCD) in the context of human-computer interaction. Inspired by these approaches, we make an alternate use of sound for semantic image segmentation, firstly to help predict the materials of everyday objects, and secondly to use these material predictions to refine our overall predictions of object categories. The use of audio data has also been beneficial in other vision contexts such as the detection of violent scenes in videos [10], object tracking [3, 8] and human action recognition [28].

The sounds that we perceive (and microphones sense) are pressure waves travelling through the air [16]. When an object is struck, particles within it vibrate. Pressure waves are then transmitted through the object (and potentially reflected as well). The object's density and volume determine the transmission and reflection of the emitted sound waves. Thus, the sound that an object emits is more dependent on the material from which it is made than its object class. By using sound, we are able to infer information about an object's material properties that would be difficult or impossible to obtain by visual means. We propose that by making use of sound to complement the existing visual data, we can capture discriminative information and achieve significant improvements in object recognition.

Approaches involving the estimation of multiple labels in images are plentiful. For example, Goldluecke *et al*. [11] used convex relaxation in order to jointly optimise for depth and occlusions in a scene. Taking advantage of the dependencies between pairs of pixel labels, several authors [14, 19, 21, 34] observed a significant improvement in object class and depth estimation in comparison to estimating them independently. Here, we consider estimating the overall category and material properties of an object, which are related and mutually beneficial. As an example, tables are often made of wood, whilst mugs are not.

Due to the success of the aforementioned approaches, we adopt a similar random field framework to [21, 34] in which to optimise a joint energy. To provide the necessary inputs to our framework, we employ similar object class features to Ladicky *et al*. [21], with additional material features derived from both auditory and visual data. Our auditory data is obtained by tapping the objects of interest with a human knuckle, and is therefore sparsely-distributed over the scene. As a result, it makes sense for us to use visual features as well in order to help predict material properties for areas of the scene for which we do not have audio. We leverage recent advances in solving densely-connected pairwise conditional random field (CRF) models [17, 34] to implement our proposed object-material multi-label inference approach.

## 3 Audio-Visual Dataset

The new dataset on which we evaluate our approach consists of 9 long reconstruction sequences[1], containing on average 1600 individual $640 \times 480$ RGB-D frames. The sequences were captured using a consumer-grade depth camera (an ASUS Xtion Pro) running at 30 FPS. We annotated these sequences using 20 different object classes and 11 material types. In the supplementary material, we show the distributions of these labels and sample images with associated ground-truth labels.

---

[1]Available at: http://www.robots.ox.ac.uk/~tvg/projects/AudioVisual/

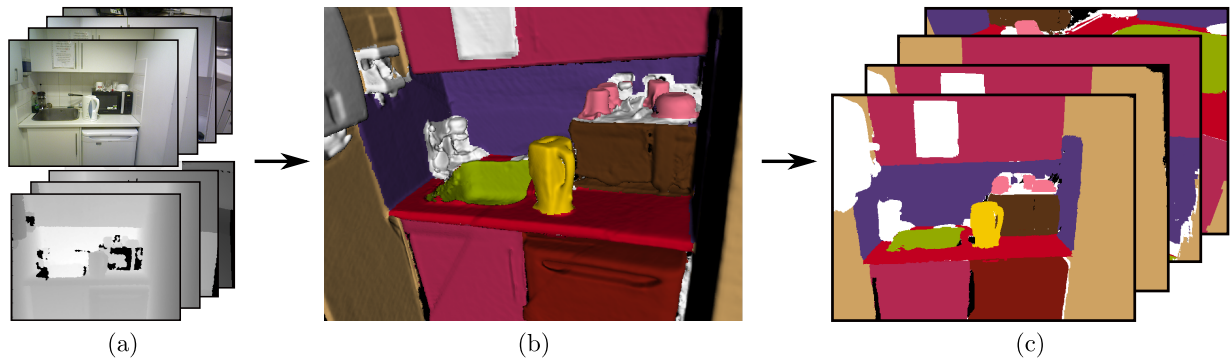

(a) 　　　　　　　　　　　(b) 　　　　　　　　　　　(c)

Figure 2: *Illustration of the labelling pipeline. (a) A sequence of colour and depth images are captured using an RGB-D camera. (b) The scene is reconstructed by integrating the depth images, and then manually labelled in 3D. Note that only object labels are shown in (b). (c) Ground-truth labellings associated with the colour images captured in (a), generated by raycasting the 3D reconstruction (b) from different viewpoints.*

Our dataset differs from other segmentation datasets [6, 7, 13, 30, 31, 36] in a number of important ways. Firstly, we provide sparse audio data that characterise the sounds emitted by objects when they are struck by a human knuckle, together with dense labels indicating both object and material categories for every pixel in the scene (by contrast, existing datasets do not provide material categories). Secondly, our dataset focuses on realistic, cluttered indoor scenes in which little to no contrast or texture is present. As illustrated by our experiments in Section 5.1, these scenes are significantly harder to segment than those in existing datasets; indeed, we carefully designed the dataset to be extremely challenging for state-of-the-art labelling techniques. Thirdly, in contrast to previous datasets, our dataset was annotated in 3D: this not only reduced the manual effort required to label the scenes, but made it easier for us to achieve consistent labellings of the scene from different viewpoints. Such consistent labellings are particularly useful for testing temporally-consistent video segmentation, as the difficulties encountered by [19, 23] in evaluating segmentation performance without having per-frame ground truth annotations on the CamVid dataset [6] will not be encountered.

**Annotation using 3D reconstructions of scenes.** In order to annotate each scene in our dataset in 3D, we first reconstructed it from a sequence of depth images (Fig. 2a), using the online 3D reconstruction system of [26, 29]. Each reconstructed scene was then manually annotated in 3D (Fig. 2b) using an interactive scene segmentation framework [12, 33]. Finally, 2D ground truth labellings of the scene (Fig. 2c) were generated for each original viewpoint by raycasting the labelled 3D scene.

This approach allowed us to forgo the time- and labour-intensive process of labelling each image individually by hand [7, 30, 31]. Typically, we were able to label an image sequence of 2000 frames in approximately 45 minutes, which compares favourably with the 20-25 minutes that was required to annotate each frame of the CamVid database by hand [6]. Moreover, most of our annotation time (35 out of 45 minutes) was spent refining the labelling of object boundaries. Although our method of labelling was limited by the available resolution of the depth sensor and was not able to capture glass objects (since the infrared light used by structured light sensors such as the Xtion is absorbed by glass), it drastically reduced the time required to obtain per-pixel annotations of the images.

**Auditory data.** In addition to capturing images of the scene, uncompressed audio was also recorded for specific scene objects using a high-quality portable condensor microphone (Samson GoMic at 44.1kHz). Note that the collected audio data can be used independently for audio classification, such as in [32]. In total, we collected approximately 600 sounds from 50 different objects and 9 material categories. The objects we struck are detailed further in the supplementary material. Due to the localised nature of sounds emitted from objects, we can only associate sound data with the points at which they were struck. We perform this by annotating the approximate location at which the object was struck in the 3D reconstruction. The exact number of pixels associated with a sound measurement in the 2D projection (yellow polygons in Fig. 1a) then depends on the viewpoint (for example, more pixels will be associated when zoomed in on the object). The median number of pixels associated with a sound is 575, which is 0.18% of the total number of pixels in the image.

**Training, validation and test folds.** Our training, validation and test folds were approximately 55%, 15% and 30% of the total data respectively. As detailed in the supplementary material, these folds were chosen so that images in the test set were not from the same scenes as the images in the training and validation sets.

## 4 Methodology

As mentioned in Section 1, estimating material properties from visual data alone can be difficult, and so it can be helpful to incorporate audio information to help distinguish between visually-similar materials. However, since audio information obtained by tapping objects is only available at sparse locations in an image, we need a way of propagating this information to the whole image. To achieve this, we introduce a densely-connected, pairwise Conditional Random Field (CRF) model that supports long-range interactions and enforces consistency between connected nodes. In practice, since estimates of object and material properties can be mutually informative, we use a two-layer CRF to model the joint estimation of object and material labels, and allow the two types of estimate to influence each other by connecting the two layers of the CRF with joint potentials.

### 4.1 Random Field Model

We model the joint estimation of object and material labels in an energy minimisation framework. Each image pixel $i \in \mathcal{V} = \{1,...,N\}$ is associated with a discrete random variable $X_i = [O_i, M_i]$ that takes a label $x_i = [o_i, m_i]$ from the product label space of object and material labels, $\mathcal{O} \times \mathcal{M}$. In our case, the object labels $\mathcal{O} = \{o_1,...,o_O\}$ correspond to different object classes such as desk, wall and floor, whilst the material labels $\mathcal{M} = \{m_1,...,m_M\}$ refer to classes such as wood, plastic and ceramic. We wish to find the best (*maximum a posteriori*) labelling $\mathbf{x}^* = [\mathbf{o}^*, \mathbf{m}^*]$ associated with the minimum energy of the two-layer CRF, expressed in its general form as

$$P(\mathbf{x}|\mathbf{D}) = \frac{1}{Z(\mathbf{D})} \exp(-E(\mathbf{x}|\mathbf{D})), \qquad E(\mathbf{x}|\mathbf{D}) = \sum_{c \in \mathcal{C}} \psi_c(\mathbf{x}_c|\mathbf{D}), \qquad (1)$$

where $E(\mathbf{x}|\mathbf{D})$ is the energy associated with labelling $\mathbf{x}$, conditioned on the visual and auditory data $\mathbf{D} = \{\mathbf{I}, \mathbf{A}\}$, $Z(\mathbf{D}) = \sum_{\mathbf{x}'} \exp(-E(\mathbf{x}'|\mathbf{D}))$ is the (data-dependent) partition function and each clique $c \in \mathcal{C}$ induces a potential $\psi_c(\cdot)$.

**Joint Model.** Since we are jointly estimating object and material labellings, we define a two-layer CRF energy function over discrete variables $\mathbf{O}$ and $\mathbf{M}$ respectively [34]. The two layers are then connected via pairwise potentials to take correlations between objects and materials into account:

$$E(\mathbf{x}|\mathbf{D}) = E^O(\mathbf{o}|\mathbf{I}) + E^M(\mathbf{m}|\mathbf{I},\mathbf{A}) + E^J(\mathbf{o},\mathbf{m}|\mathbf{I},\mathbf{A}) \quad (2)$$

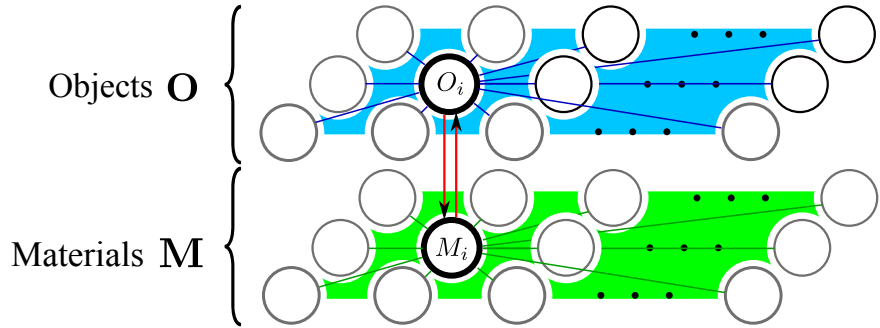

**Figure 3:** *Model of our Random Field formulation, focusing on nodes $O_i$ and $M_i$. The variables, **O** and **M**, corresponding to objects and materials respectively, are modelled as separate layers of a fully-connected CRF with pairwise terms between each corresponding node $O_i$ and $M_i$.*

In this, $E^O(\mathbf{o}|\mathbf{I})$ is the energy for the object labelling, conditioned on image data **I**, $E^M(\mathbf{m}|\mathbf{I},\mathbf{A})$ is the energy for the material labelling, conditioned on image data **I** and audio data **A**, and $E^J(\mathbf{o},\mathbf{m}|\mathbf{I},\mathbf{A})$ is the joint energy function linking the object and material domains. Without the joint term in Equation 2, the energy would decompose into two, i.e. we would effectively be solving an object CRF and a material CRF in parallel with no links between them (Fig. 3).

**Object labelling.** For object segmentation, we use a formulation with unary and densely-connected pairwise terms allowing long-range interactions [17]

$$E^O(\mathbf{o}|\mathbf{I}) = \sum_{i \in \mathcal{V}} \psi_u^O(o_i) + \sum_{i < j \in \mathcal{V}} \psi_p^O(o_i, o_j) \quad (3)$$

In this, the unary potential terms $\psi_u^O(\cdot)$, implicitly conditioned on image data **I**, correspond to the cost of pixel $i$ taking an object label $o_i \in \mathcal{O}$, as described in Section 4.2. The pairwise potential function $\psi_p^O(\cdot,\cdot)$ enforces consistency between the connected variables and takes the form of a mixture of Gaussian kernels as in [17], allowing efficient mean-field inference in densely-connected models. Like [17], the Gaussian kernel parameters are obtained using cross-validation.

**Materials.** For materials, we use a similar energy function:

$$E^M(\mathbf{m}|\mathbf{I},\mathbf{A}) = \sum_{i \in \mathcal{V}} \psi_u^M(m_i) + \sum_{i < j \in \mathcal{V}} \psi_p^M(m_i, m_j) \quad (4)$$

In this, the unary $\psi_u^M(\cdot)$ and pairwise $\psi_p^M(\cdot,\cdot)$ potentials are implicitly conditioned on both image data **I** and audio data **A**. As in the case of object labelling, our pairwise material potential $\psi_p^M(\cdot,\cdot)$ takes the form of a mixture of Gaussian kernels. Our material unary is defined to be the negative logarithm of a convex combination of visual and auditory terms:

$$\psi_u^M(m_i) = \begin{cases} -\ln\left[w_{av}p(m_i|\mathbf{I}) + (1-w_{av})p(m_i|\mathbf{A})\right] & \text{if audio data is present} \\ -\ln\left[w_v p(m_i|\mathbf{I}) + (1-w_v)U\right] & \text{otherwise,} \end{cases} \quad (5)$$

where our visual term is defined to be $p(m_i|\mathbf{I})$ for each pixel $i$, whilst our auditory term is defined to be $p(m_i|\mathbf{A})$ for the sparse set of pixels $i$ for which audio information is available, and a uniform distribution $U$ otherwise. The weights $w_{av}, w_v \in [0,1]$ control the balance between the two terms for pixels for which audio information is and is not available, respectively.

The justification for using a uniform auditory term when no sound information is available (rather than simply setting $w_v$ to 1 and relying on the available visual information) is that it allows us to ameliorate any poor predictions made by the visual classifier. In particular, it is possible for the visual classifier to confidently predict an incorrect material class for some pixels (see Section 5), which can inhibit the propagation of labels predicted by the sound classifier through the image. By introducing a uniform auditory term, we can lessen the influence of the visual classifier in such situations and allow the labels to propagate more effectively.

**Joint potentials.** The energy $E^J(\mathbf{o}, \mathbf{m} | \mathbf{I}, \mathbf{A}) = \sum_i \psi_p^J(o_i, m_i)$ captures the correlation between the object and material labels (Fig. 3). The joint potential $\psi_p^J(\cdot, \cdot)$ connects random variables in the two CRF layers [34], and is defined as the negative logarithm of the conditional distribution of object and material labels observed from the training data:

$$\psi_p^J(o_i, m_i) = -w_{mo} \ln\left(p(o_i | m_i)\right) - w_{om} \ln\left(p(m_i | o_i)\right). \tag{6}$$

**Inference.** Given the energy function in Equation 2, and the form of the pairwise potentials just described, we use an efficient filter-based variant of mean-field optimisation [1, 17] to infer the optimal joint assignment of object and material labels. The mean-field update equations for our CRF model can be found in the supplementary material.

## 4.2 Unary potentials

The per-pixel visual unary potentials were obtained by training a joint boosting classifier on visual features, whilst auditory unary potentials were obtained by training a random forest classifier on auditory features. The outputs of the two classifiers are probability distributions over object and material categories, respectively, and these were converted to energies by taking their negative logarithms.

**Visual features.** For our per-pixel visual features, we used the 17-dimensional Texton-Boost filter bank of [30], adding colour, SIFT and LBP features in a manner similar to [20].

**Auditory features.** The sound recordings require windowing before classification, since typically, a recording begins before the object is struck and ends after the sound has decayed. To do this, we subdivided the input waveform into sub-windows of size $l$, and selected the $k$ consecutive windows with the highest $\ell_2$ norm, as shown in Fig. 4. We rely on the assumption that the sound of interest initially has the highest amplitude in the recording and then decays over time. Optimal values of the integers $l$ and $k$ were determined by cross-validation.

From the isolated sound, we extracted the following features: energy per window, energy entropy [10], zero crossing rate [9], spectral centroid, spread, flux, rolloff and entropy [2, 10]. The first three are computed in the time domain whilst the remainder are evaluated from the spectra computed by the Short-Time Fourier Transform [27].

## 5 Experimental evaluation

In order to quantify the performance of our system, we evaluate it on the new dataset we presented in Section 3 using the following performance measures: accuracy (the percentage of pixels labelled correctly), intersection-over-union score (as used in the PASCAL segmentation challenge [7]) and F1-score (the harmonic mean of precision and recall [7]).

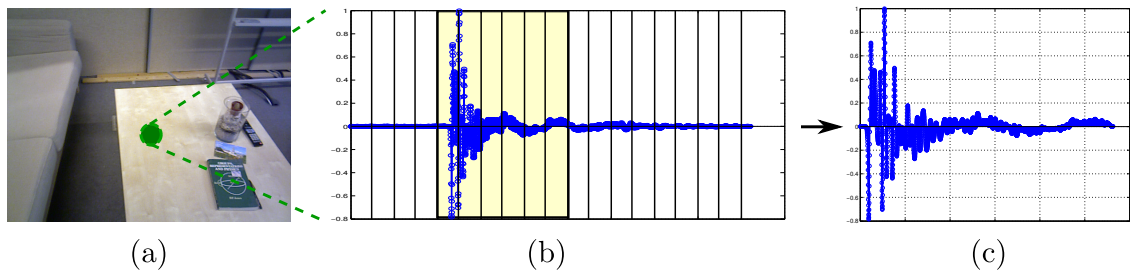

Figure 4: *Segmenting the sound of a wooden table being struck* **(a)** *The location where the table was struck.* **(b)** *The recording of the waveform (with background noise before and after the actual contact), subdivided into windows of $l = 512$ samples.* **(c)** *The sound segmented from the $k = 30$ consecutive windows with the highest $\ell_2$ norm (fewer windows have been shown for illustrative purposes).*

Table 1: Per-class quantitative performance of the unary sound classifier

| Material | Plastic | Wood | Gypsum | Ceramic | Melamine | Tile | Steel | Cotton | Average |
|---|---|---|---|---|---|---|---|---|---|
| Accuracy (%) | 73.61 | 100 | 16.67 | 100 | 33.33 | 14.29 | 11.11 | 0 | 67.11 |
| F1-Score (%) | 82.81 | 59.52 | 16.67 | 97.30 | 42.86 | 20 | 20 | 0 | 42.39 |

First, we evaluate the performance of our unary sound classifier to identify material categories likely to benefit from tapping. Next, we show the performance achieved when using visual features to label object and material classes independently. Finally, we demonstrate how incorporating auditory features can improve our predictions of material categories, and the effect this has on improving the recognition of object classes after joint optimisation.

## 5.1 Quantitative results and discussion

**Unary sound classifier performance.** As seen in Table 1, our unary sound classifier predicts certain material classes (wood, ceramic, plastic) with high accuracy, since these materials produce distinctive sounds.

However, our sound classifier does not achieve perfect results, as the sound emitted from an object is also determined by its density, volume and the pressure being applied to it [16]. We also observed the adverse effect that sound transmission has on recognition, e.g. knocking a tile affixed to a wall causes the resultant pressure waves in the tile to propagate through the wall. The end result is a sound wave that sounds similar to the wall and distinct from the sound of knocking a tile placed on a wooden table. Striking a melamine whiteboard that is affixed to a wall produces the same effect. Finally, striking objects such as cotton chairs does not produce enough sound to accurately identify them.

**Using visual features alone.** Table 2 shows the accuracy in labelling objects and materials, first using only visual features, and then introducing auditory cues as well. The performance measures we obtain from training our pixel classifiers using only visual features, and a CRF with unary and pairwise terms (third row), serves as the baseline for when we introduce sound features. We consider our dataset to be a challenging one given that our cluttered indoor scenes do not show much variation in texture or colour between different objects (as shown in Fig. 1), making them difficult to recognise. We confirm this by testing the CRF implementation of [20] (which has pairwise terms between pixels in an 8-neighbourhood) [2] which has a similarly low accuracy when using the same unaries. However, our mean-field based inference is significantly faster.

---

[2]Source code available at: http://www.inf.ethz.ch/personal/ladickyl/ALE.zip

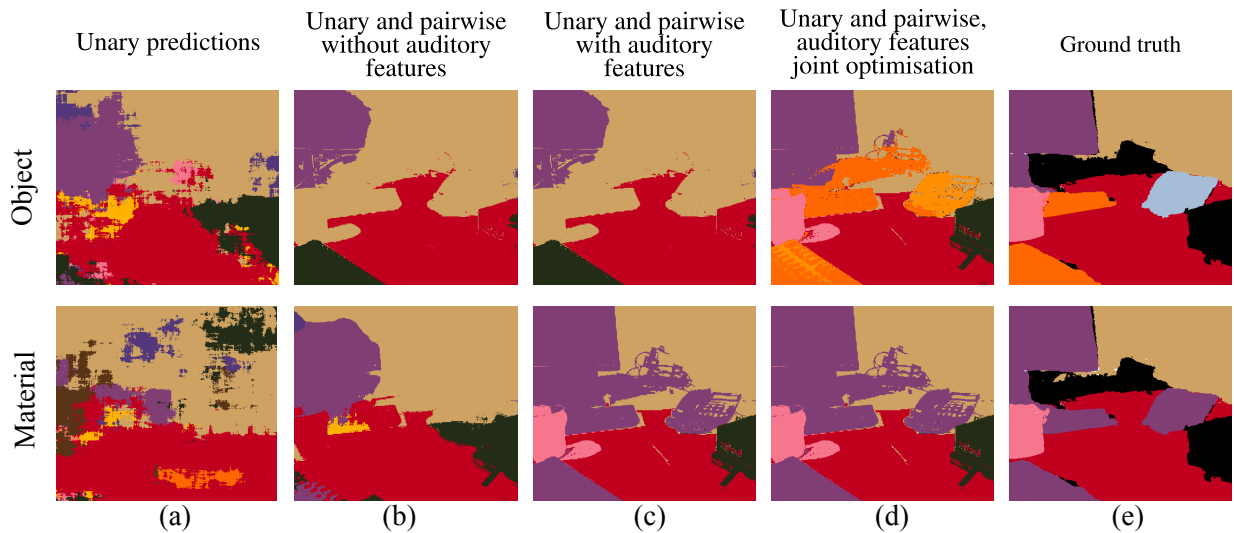

Figure 5: *(a) The noisy predictions made by the per-pixel unary classifiers. The indicated regions in the input image show the locations where sound information is present. (b) The output of the CRF using only visual features. (c) The material unaries from the sound classifier increases the probability of the demarcated pixels taking the labels "plastic" and "ceramic" such that it is enough to propogate throughout the keyboard, telephone and mug. Note that long-range interactions present in our densely-conneced CRF model which enables the monitor and upper keyboard to take on material labels even though they have no sound information in this image. (d) Finally, the joint optimisation between material and object labels results in the keyboard and mug taking the correct label. The telephone is now misclassified as "mouse," but both these objects are made of plastic. (e) Note, limitations on our 3D reconstruction constrain the resolution of the labelling. The black, "void", labels are ignored when computing performance metrics.*

Table 2: Results of semantic segmentation of object and material labels

|  | Weighted Mean IoU | | Mean IoU | | Accuracy(%) | | Mean F1-Score | |
|---|---|---|---|---|---|---|---|---|
|  | Object | Material | Object | Material | Object | Material | Object | Material |
| Visual features only (unary) | 31.51 | 38.97 | 10.16 | 16.71 | 49.89 | 58.46 | 15.54 | 25.00 |
| Visual features only (unary and pairwise) [20] | 32.54 | 40.20 | 10.69 | 17.09 | 52.19 | 60.81 | 16.06 | 25.28 |
| Visual features only (unary and pairwise) | 32.64 | 41.06 | 10.88 | 17.65 | 52.84 | 62.46 | 16.15 | 25.91 |
| Audiovisual features (unary and pairwise) | – | 44.54 | – | 21.83 | – | 66.45 | – | 31.49 |
| Visual features only, joint inference | 34.40 | 41.06 | 11.15 | 17.65 | 53.63 | 62.46 | 17.19 | 25.91 |
| Audiovisual features, joint inference | 36.79 | 44.54 | 12.80 | 21.83 | 55.65 | 66.45 | 19.59 | 31.49 |

**The addition of auditory features.** Adding sound features when classifying materials improves the weighted mean intersection-over-union (IoU) by 3.5%. The F1-score and accuracy increase by 5.6% and 4% respectively. At this stage, object-classification performance is unchanged, since auditory features are used for material identification only. We have seen from Table 1 that the unary sound classifier predicts a number of material categories with high accuracy. Moreover, as shown in Fig. 5c, our random-field model propagates the unary sound classifier's prediction throughout the rest of the object. The pairwise potentials, which encourage nearby pixels of similar appearance to take on the same label, and unary potentials from the visual features, govern this propagation.

**Joint optimisation of object and material labels.** Next, we use the improved material labelling to also improve the object labelling. This is achieved through the joint potential terms in Equation 2. As shown in Fig. 5d, we use the fact that object and material classes are correlated in order to impose consistency between these labels (for example, mugs are made of ceramic, but not of gypsum). Through cross-validation, we found that the best results were achieved when the cost from object-to-material was very low ($w_{om} = 0.05$ and $w_{mo} = 3$ from Equation 6). This was because our material predictions are stronger than our object ones, and hence, the object nodes are more likely to send incorrect information to the material nodes in the CRF. Joint optimisation produces better object labelling even when we do not use any sound features to improve material labelling as shown in the penultimate row of our table. However, augmenting our material classification with auditory features and then performing joint inference produces the best results. As Fig. 5 shows, objects which are made of ceramic, plastic or wood show particular improvement, since these are the material categories that the unary sound classifier was most accurate at predicting from Table 1.

## 6 Conclusions and Future Work

In this paper, we have demonstrated how complementary sensory modalities can be used to improve classification performance. In particular, we have shown that by using a Conditional Random Field model, we can use sparse auditory information effectively to augment existing visual data and help better predict the material properties of objects. We have further shown that by exploiting the correlation between material and object classes, it is possible to use these better material predictions to achieve better object category classification.

To facilitate our experiments in this paper, we have produced a new RGB-D dataset that provides dense per-pixel object and material labels for a variety of indoor scenes that contain minimal contrast or texture. Our dataset also provides auditory data obtained from striking various objects in the scene at sparse locations. We hope that our challenging dataset will encourage future work in audio-visual semantic segmentation.

In terms of further work, we believe that investigating the use of other complementary sensory modalities in addition to sound could yield further improvements in classification performance. Moreover, we aim to implement our system on a mobile robot which will use its robotic arm to tap objects, record the resulting sound, and learn more about its environment.

## References

[1] Andrew Adams, Jongmin Baek, and Myers Abraham Davis. Fast high-dimensional filtering using the permutohedral lattice. In *Computer Graphics Forum*, 2010.

[2] Fabio Antonacci, Giorgio Prandi, Giuseppe Bernasconi, Roberto Galli, and Augusto Sarti. Audio-based object recognition system for tangible acoustic interfaces. In *Haptic Audio visual Environments and Games, 2009. HAVE 2009. IEEE International Workshop on*, 2009.

[3] Matthew J Beal, Nebojsa Jojic, and Hagai Attias. A graphical model for audiovisual object tracking. *T-PAMI*, 2003.

[4] Jeannette Bohg, Antonio Morales, Tamim Asfour, and Danica Kragic. Data-driven grasp synthesis - A survey. *CoRR*, 2013.

[5] Xavier Boix, Josep M Gonfaus, Joost van de Weijer, Andrew D Bagdanov, Joan Serrat, and Jordi Gonzàlez. Harmony potentials. *IJCV*, 2012.

[6] Gabriel J Brostow, Julien Fauqueur, and Roberto Cipolla. Semantic object classes in video: A high-definition ground truth database. *Pattern Recognition Letters*, 2009.

[7] Mark Everingham, Luc Van Gool, Christopher KI Williams, John Winn, and Andrew Zisserman. The pascal visual object classes (voc) challenge. *IJCV*, 2010.

[8] Daniel Gatica-Perez, Guillaume Lathoud, Jean-Marc Odobez, and Iain McCowan. Audiovisual probabilistic tracking of multiple speakers in meetings. *Audio, Speech, and Language Processing, IEEE Transactions on*, 2007.

[9] T. Giannakopoulos, A. Pikrakis, and S. Theodoridis. A multi-class audio classification method with respect to violent content in movies using bayesian networks. In *Multimedia Signal Processing, 2007. MMSP 2007. IEEE 9th Workshop on*, Oct 2007.

[10] Theodoros Giannakopoulos, Alexandros Makris, Dimitrios Kosmopoulos, Stavros Perantonis, and Sergios Theodoridis. Audio-visual fusion for detecting violent scenes in videos. In *Artificial Intelligence: Theories, Models and Applications*. 2010.

[11] Bastian Goldluecke and Daniel Cremers. Convex relaxation for multilabel problems with product label spaces. In *ECCV*, 2010.

[12] Stuart Golodetz, Michael Sapienza, Julien P C Valentin, Vibhav Vineet, Ming-Ming Cheng, Victor A Prisacariu, Olaf Kähler, Carl Yuheng Ren, Anurag Arnab, Stephen L Hicks, David W Murray, Shahram Izadi, and Philip H S Torr. SemanticPaint: Interactive Segmentation and Learning of 3D Worlds. Demo in SIGGRAPH ET, 2015.

[13] Stephen Gould, Richard Fulton, and Daphne Koller. Decomposing a Scene into Geometric and Semantically Consistent Regions. In *ICCV*, 2009.

[14] Christian Häne, Christopher Zach, Andrea Cohen, Roland Angst, and Marc Pollefeys. Joint 3D scene reconstruction and class segmentation. In *CVPR*, 2013.

[15] Chris Harrison, Desney Tan, and Dan Morris. Skinput: Appropriating the body as an input surface. In *CHI*, 2010.

[16] Y.H. Kim. *Sound Propagation: An Impedance Based Approach*. Wiley, 2010.

[17] P. Krähenbühl and V. Koltun. Efficient inference in fully connected CRFs with Gaussian edge potentials. In *NIPS*, 2011.

[18] Christian Kroos and Katherine Hogan. Visual influence on auditory perception: Is speech special? In *International Conference on Audio-Visual Speech Processing*, 2009.

[19] Abhijit Kundu, Yin Li, Frank Daellert, Fuxin Li, and James M. Rehg. Joint semantic segmentation and 3D reconstruction from monocular video. In *ECCV*, 2014.

[20] Lubor Ladicky, Christopher Russell, Pushmeet Kohli, and Philip HS Torr. Associative hierarchical crfs for object class image segmentation. In *ICCV*, 2009.

[21] Lubor Ladicky, Paul Sturgess, Chris Russell, Sunando Sengupta, Yalin Bastanlar, William Clocksin, and Philip H.S. Torr. Joint optimization for object class segmentation and dense stereo reconstruction. *IJCV*, 2012.

[22] Victor Lempitsky, Andrea Vedaldi, and Andrew Zisserman. Pylon model for semantic segmentation. In *NIPS*, 2011.

[23] Ondrej Miksik, Daniel Munoz, J Andrew Bagnell, and Martial Hebert. Efficient temporal consistency for streaming video scene analysis. In *ICRA*, 2013.

[24] Ondrej Miksik, Yousef Amar, Vibhav Vineet, Patrick Perez, and Philip H. S. Torr. Incremental dense multi-modal 3d scene reconstruction. In *IROS*, 2015.

[25] Ondrej Miksik, Vibhav Vineet, Morten Lidegaard, Ram Prasaath, Matthias Nießner, Stuart Golodetz, Stephen L. Hicks, Patrick Perez, Shahram Izadi, and Philip H. S. Torr. The Semantic Paintbrush: Interactive 3D Mapping and Recognition in Large Outdoor Spaces. In *CHI*. ACM, 2015.

[26] M. Nießner, M. Zollhöfer, S. Izadi, and M. Stamminger. Real-time 3d reconstruction at scale using voxel hashing. *ACM Transactions on Graphics (TOG)*, 2013.

[27] Alan V. Oppenheim, Ronald W. Schafer, and John R. Buck. *Discrete-time Signal Processing (2nd Ed.)*. Prentice-Hall, Inc., Upper Saddle River, NJ, USA, 1999.

[28] Alessandro Pieropan, Giampiero Salvi, Karl Pauwels, and Hedvig Kjellstrom. Audio-visual classification and detection of human manipulation actions. In *IROS*, 2014.

[29] V. A. Prisacariu, O. Kähler, M. M. Cheng, C. Y. Ren, J. Valentin, P. H. S. Torr, I. D. Reid, and D. W. Murray. A Framework for the Volumetric Integration of Depth Images. *ArXiv e-prints*, 2014.

[30] Jamie Shotton, John Winn, Carsten Rother, and Antonio Criminisi. Textonboost for image understanding: Multi-class object recognition and segmentation by jointly modeling texture, layout, and context. *IJCV*, 2009.

[31] Nathan Silberman, Derek Hoiem, Pushmeet Kohli, and Rob Fergus. Indoor Segmentation and Support Inference from RGBD Images. In *ECCV*, 2012.

[32] Jivko Sinapov, Mark Wiemer, and Alexander Stoytchev. Interactive learning of the acoustic properties of household objects. In *ICRA*, 2009.

[33] Julien P C Valentin, Vibhav Vineet, Ming-Ming Cheng, David Kim, Jamie Shotton, Pushmeet Kohli, Matthias Nießner, Antonio Criminisi, Shahram Izadi, and Philip H S Torr. SemanticPaint: Interactive 3D Labeling and Learning at your Fingertips. 2015.

[34] Vibhav Vineet, Jonathan Warrell, and Philip HS Torr. Filter-based mean-field inference for random fields with higher-order terms and product label-spaces. *IJCV*, 2014.

[35] Vibhav Vineet, Ondrej Miksik, Morten Lidegaard, Matthias Nießner, Stuart Golodetz, Victor A. Prisacariu, Olaf Kahler, David W. Murray, Shahram Izadi, Patrick Perez, and Philip H. S. Torr. Incremental dense semantic stereo fusion for large-scale semantic scene reconstruction. In *ICRA*, 2015.

[36] Christian Wojek and Bernt Schiele. A Dynamic Conditional Random Field Model for Joint Labeling of Object and Scene Classes. In *ECCV*, 2008.

[37] Jian Yao, Sanja Fidler, and Raquel Urtasun. Describing the Scene as a Whole: Joint Object Detection, Scene Classification and Semantic Segmentation. In *CVPR*, 2012.

[38] Shuai Zheng, Ming-Ming Cheng, Jonathan Warrell, Paul Sturgess, Vibhav Vineet, Carsten Rother, and Philip H. S. Torr. Dense semantic image segmentation with objects and attributes. In *CVPR*, 2014.